\documentclass[10pt,a4paper,conference]{IEEEtran}

\usepackage{times}
\usepackage{epsfig}
\usepackage{graphicx}
\usepackage{amsmath}
\usepackage{amssymb}
\usepackage{float}
\usepackage{nicefrac}
\usepackage{tabularx}
\usepackage{booktabs}
\usepackage{multirow}
\usepackage{makecell}

\usepackage[pagebackref=true,breaklinks=true,bookmarks=false]{hyperref}

\begin{document}

%%%%%%%%% TITLE
%\title{Deep learning for weakly labeled anomaly detection}
\title{End-to-end training of a two-stage neural network for defect detection}
%\title{Enabling weakly labeled anomaly detection with end-to-end learning for deep architecture}

%\author{\IEEEauthorblockN{Michael Shell}
%\IEEEauthorblockA{School of Electrical and\\Computer Engineering\\
%Georgia Institute of Technology\\
%Atlanta, Georgia 30332--0250\\
%Email: http://www.michaelshell.org/contact.html}}

\author{\IEEEauthorblockN{Jakob Bo\v{z}i\v{c}, Domen Tabernik and Danijel Sko\v{c}aj}\\
\IEEEauthorblockA{University of Ljubljana, Faculty of Computer and Information Science\\
Ve\v{c}na pot 113, 1000 Ljubljana\\
{\tt\small jb3065@student.uni-lj.si, domen.tabernik@fri.uni-lj.si}, \\ 
{\tt\small danijel.skocaj@fri.uni-lj.si}
}
}

\maketitle

%%%%%%%%% ABSTRACT
\begin{abstract}
    %Deep-learning methods are nowadays often used to address the surface defect detection problems for the industrial quality control. However, with a large amount of data needed for learning, often requiring high-precision labels, many industrial problems cannot be easily solved or the cost of solutions significantly increases due to the annotation requirements. In this work, we explore the use of weakly labeled data for surface defect detection to reduce the need for the highly detailed annotations. We propose an end-to-end learning for a two-stage network that enables mixing of fully and weakly annotated data during the training. We demonstrate on the KolektorSDD dataset, that the proposed model outperforms the recently proposed state-of-the-art method and we demonstrate on DAGM and Severstal Steel Defect datasets that the same model requires a significantly smaller amount of fully annotated data to achieve state-of-the-art results.

    Segmentation-based, two-stage neural network has shown excellent results in the surface defect detection, enabling the network to learn from a relatively small number of samples.
    In this work, we introduce end-to-end training of the two-stage network together with several extensions to the training process, which reduce the amount of training time and improve the results on the surface defect detection tasks.
    To enable end-to-end training we carefully balance the contributions of both the segmentation and the classification loss throughout the learning.
    We adjust the gradient flow from the classification into the segmentation network in order to prevent the unstable features from corrupting the learning.
    As an additional extension to the learning, we propose frequency-of-use sampling scheme of negative samples to address the issue of over- and under-sampling of images during the training, while we employ the distance transform algorithm on the region-based segmentation masks as weights for positive pixels, giving greater importance to areas with higher probability of presence of defect without requiring a detailed annotation.
    We demonstrate the performance of the end-to-end training scheme and the proposed extensions on three defect detection datasets---DAGM, KolektorSDD and Severstal Steel defect dataset--- where we show state-of-the-art results. On the DAGM and the KolektorSDD we demonstrate 100\% detection rate, therefore completely solving the datasets.
    Additional ablation study performed on all three datasets quantitatively demonstrates the contribution to the overall result improvements for each of the proposed extensions.

\end{abstract}

\IEEEpeerreviewmaketitle

%%%%%%%%% BODY TEXT

\section{Introduction}

Quality inspection of the produced items and their surfaces is a very important part of the industrial production processes. Traditionally, classical machine-vision methods have been in use to automate the visual quality inspection processes, however with the introduction of the Industry 4.0 paradigm, newer deep-learning based algorithms have started being employed~\cite{Weimer2016,Racki2018,Lin2018,Tabernik2019JIM}. Deep-learning models have become suitable for this task due to their larger capacity to model the complex features and easier adaptation to the different products without the explicit feature hand-engineering. 

A novel two-stage architecture~\cite{Racki2018,Tabernik2019JIM,Xu2020} has proven highly successful in the surface defect detection. State-of-the-art performance can be contributed to a two-stage design, where a defect segmentation is performed in the first stage, followed by a per-image classification on defective vs. non-defective surfaces in the second stage. This enables learning with only a small number of positive training samples available, as demonstrated in our previous work~\cite{Tabernik2019JIM}. The method was shown to outperform the related approaches, including a state-of-the-art commercial product. A similar two-stage architecture was also utilized by Ra\v{c}ki et al.~\cite{Racki2018} on the DAGM~\cite{Weimer2016} dataset, while Xu et al.~\cite{Xu2020} extended the two-stage architecture to a weakly supervised learning, reducing the need for the pixel-level annotations, but at a slightly compromised results. However, existing two-stage architectures rely on a cumbersome training procedure, since they require the training of the segmentation layers first, then freezing the segmentation layers and learning the classification layers afterwards. Although this two-stage learning approach produces state-of-the-art results, it also results in a slow learning process since multiple learning passes are needed.

In this work, we address the drawbacks of the two-stage architecture for the surface defect detection~\cite{Tabernik2019JIM} and propose an end-to-end training scheme that requires less precise pixel-level annotations without compromising the performance. We propose to improve the learning process by introducing the simultaneous learning of the segmentation and the classification layers in an end-to-end manner. The proposed architecture results not only in easier and faster learning of the network, but also in the improved defect detection rate. To achieve this, we propose a gradient-flow adjustment considering the pixel-level annotations. However, since the precise pixel-level annotations are difficult to obtain, we propose to use the less precise annotations that are easier to obtain. We do not utilize weakly supervised learning with only image-level tags that results in slightly lower performance, as did Xu et al.~\cite{Xu2020}, but instead extend the loss function to account for the uncertainties of the region based annotations, which allows for  significantly coarser annotations that are still fairly easy to obtain. In addition, we also introduce a frequency-of-use sampling of the non-defective samples, which even further improves the defect detection performance. We demonstrate the use of the proposed improvements on our previous implementation of the two-stage architecture~\cite{Tabernik2019JIM}, however, the same improvements are general and can be applied to any other two-stage network architecture, such as the one proposed by Xu et al.~\cite{Xu2020}. All proposed contributions are extensively evaluated in a detailed ablation study demonstrating the usefulness of the each individual component.

\begin{figure*}
    \centering
    \includegraphics[width=\linewidth]{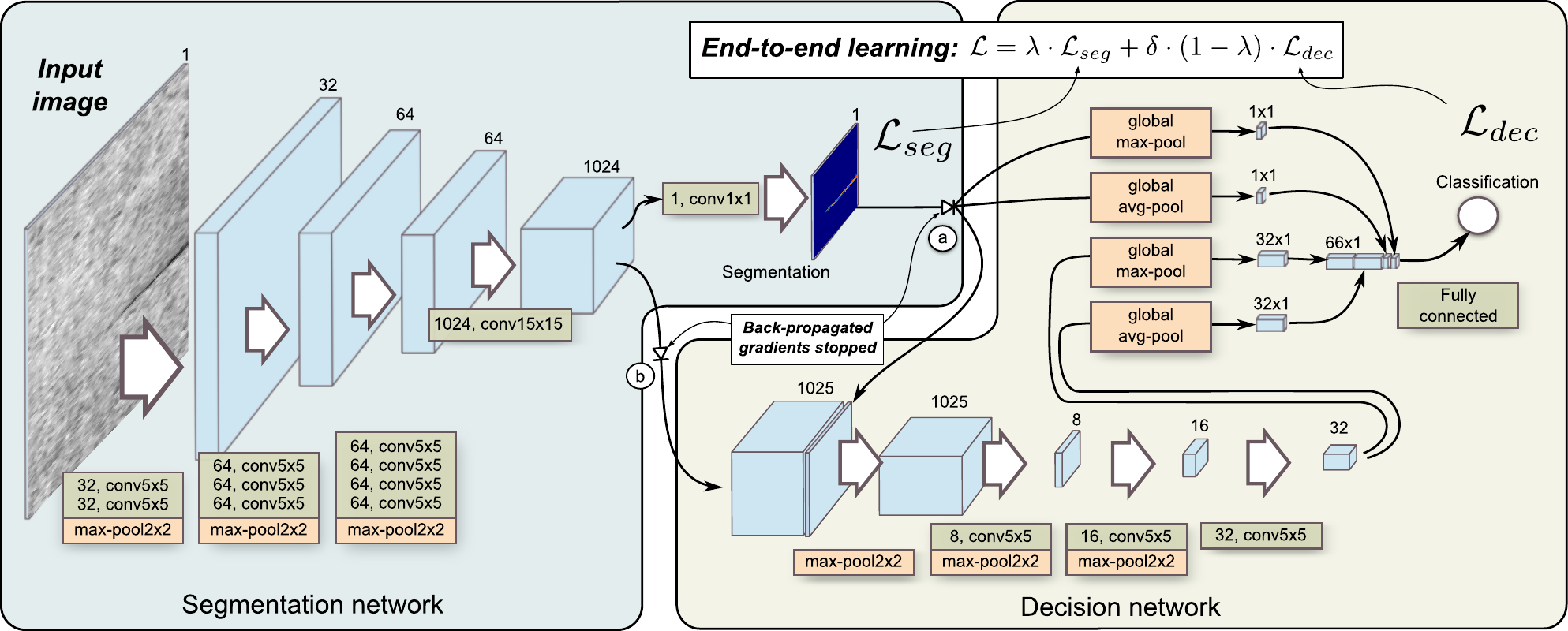}
    
    \caption{A two-stage network architecture with back-propagation changes, marked as (a) and (b), that enable end-to-end learning.} 
  \label{fig:arch}
\end{figure*}

The proposed end-to-end learning approach is evaluated on several publicly available defect detection datasets: KolektorSDD~\cite{Tabernik2019JIM}, DAGM~\cite{Weimer2016} and Severstal Steel defect dataset~\cite{SeverstalSteel2019}, demonstrating a general applicability of this approach to a multitude of defect domains. Moreover, the proposed method is shown to achieve a 100\% detection rate on DAGM~\cite{Weimer2016} and KolektorSSD, thus outperforming all other previously proposed approaches and completely solving those datasets.

The remainder of this paper is structured as follows: In Section~\ref{sec:related-work}, we present the related work, while we detail the proposed end-to-end learning with additional improvements in Section~\ref{sec:method}. In Section~\ref{sec:experiments}, we present an extensive comparison of the proposed method on three datasets, including a detailed ablation study, while we conclude with a discussion in Section~\ref{sec:conclusion}.

\section{Related work} \label{sec:related-work}

Several related works explored the use of deep-learning for the industrial anomaly detection and recognition~\cite{Masci2012,Weimer2016,Racki2018,Lin2018,Tabernik2019JIM}. The earliest work by Masci et al.~\cite{Masci2012} showed promising results when using a shallow network to address a supervised steel defect classification. A more comprehensive study of a modern deep network architecture was performed by Weimer et al.~\cite{Weimer2016}. They evaluated a number of deep-learning architectures with varying depths of layers on 6 different types of synthetic errors and showed a deep network outperforming any classical method on a synthetic dataset. However, they implemented the networks by processing individual patches of images instead of using fully convolutional approach.

Ra\v{c}ki et al.~\cite{Racki2018} proposed to improve the inefficiency of the patch-based processing from~\cite{Weimer2016} with a fully convolutional architecture. Moreover, they first utilized a two-stage architecture design that improved the performance on a synthetic dataset using the segmentation network for pixel-wise localization of the error and the decision network for per-image classification of anomalous and non-anomalous images. Our more recent work~\cite{Tabernik2019JIM} performed an extensive study of a two-stage architecture with several proposed improvements and showed that the two-stage design achieves state-of-the-art results and outperforms other architectures such as U-Net~\cite{Ronneberger2015} and DeepLab v3~\cite{Chen2017a} on a real-world case of an anomaly detection problem.

The majority of the anomaly detection papers focus on fully supervised learning with pixel-level labels. The work 
by Lin et al.~\cite{Lin2018} partially introduces weakly labeled supervision for the anomaly detection. They propose a network for detection of defects in LED chips based on the AlexNet architecture and class activation maps (CAM)~\cite{Zhou2015}. Their network allows for weakly supervised learning using only image-level labels and CAM for localizing the defects, however, they do not consider the pixel-level labels in the learning process, failing to utilize this information when available. Our previous work~\cite{Tabernik2019JIM} was also extended for weakly supervised learning by Xu et al.~\cite{Xu2020}. They used a similar two-stage architecture, but proposed a novel loss for the segmentation network enabling the learning without the pixel-level labels. However, to achieve this they proposed a three-stage learning procedure, where each network was trained separately first and additionally fine-tuned together in the end. This resulted in state-of-the-art results for weakly supervised learning but did not outperform fully labeled learning and also retained a cumbersome learning procedure that requires several stages of learning.

\section{End-to-end learning for two-stage architecture} \label{sec:method}

In this section, we present an end-to-end learning for a two-stage architecture that enables simultaneous learning of the segmentation and the classification (i.e. decision in ~\cite{Tabernik2019JIM}) layers. The overall two-stage network architecture is depicted in Figure~\ref{fig:arch}, while for more details the reader is referred to~\cite{Tabernik2019JIM}. We also present additional improvements that further increase the classification accuracy on the defect detection tasks.

\subsection{End-to-end learning}

To implement end-to-end learning, we combine both losses, the segmentation loss and the classification loss, into a single unified loss, allowing for a simultaneous learning. New combined loss is defined as:
\begin{equation} \label{eq:loss_total}
    \mathcal{L}_{total} = \lambda \cdot \mathcal{L}_{seg} + \delta \cdot (1-\lambda) \cdot \mathcal{L}_{cls},
\end{equation}
where $\mathcal{L}_{seg}$ and $\mathcal{L}_{cls}$ represent segmentation and classification losses, respectively, $\delta$ is an additional classification loss weight that prevents the classification loss from dominating the total loss, while $\lambda$ is a mixing factor that % sum to one ($\lambda_1 + \lambda_2 = 1$) and 
balances the contribution of each network in the final loss. Note that $\lambda$, and $\delta$ do not replace the learning rate $\eta$ in SGD, but complement it. They adequately control the learning process, as losses can be in different scales. The segmentation loss is averaged over all the pixels, and more importantly, only a handful of pixels are anomalous, resulting in a relatively small values compared to the classification loss, therefore we normally use smaller $\delta$ values to prevent the classification loss from dominating the total loss. 

%We also observed that classification network could not learn correctly without good segmentation features. To alleviate this issue, 
Learning the classification network before the segmentation network features are stable represents an additional challenge for the simultaneous learning. We address this by proposing to start learning only the segmentation network at the beginning and gradually progress towards learning only the classification part at the end. We formulate this by computing segmentation and classification mixing factors as a simple linear function:
\begin{align}
    \lambda &= 1 - \frac{n}{total\_epoch} 
    %\lambda_2 &= \frac{n}{total\_epoch}
\end{align}
where $n$ represents the index of the current epoch and $total\_epoch$ represents the total number of training epochs. Without the gradual mixing of both losses, the learning would in some cases result in exploding gradients, thus making the model more difficult to use. We term the process of gradually including classification network and excluding segmentation network as a \textit{dynamically balanced loss}. Additionally, using lower $\delta$ values further reduces the issue of learning on the noisy segmentation features early on, whereas using greater values has sometimes lead to the problem of exploding gradients.% in our case.
%we explored different functions that gradually increase and decrease each loss, but a simple linear function has proven as the most successful. Therefore, we use the following values:
%\begin{align}
%    \lambda_1 &= 1 - \frac{n}{total\_epoch} \\
%    \lambda_2 &= \delta\cdot\frac{n}{total\_epoch}
%\end{align}
%where $n$ represents the index of the current epoch, $total\_epoch$ represents the total number of training epochs and $\delta$ is an additional factor that prevents the classification loss from dominating the total loss. Lowering $\lambda_2$ with low $\delta$ values further reduces the issue of learning on noisy segmentation features early on, while greater $\lambda_2$ values have often lead to the problem of exploding gradients.% in our case.

\begin{figure}%[h]
    \begin{center}
    \includegraphics[width=0.45\textwidth]{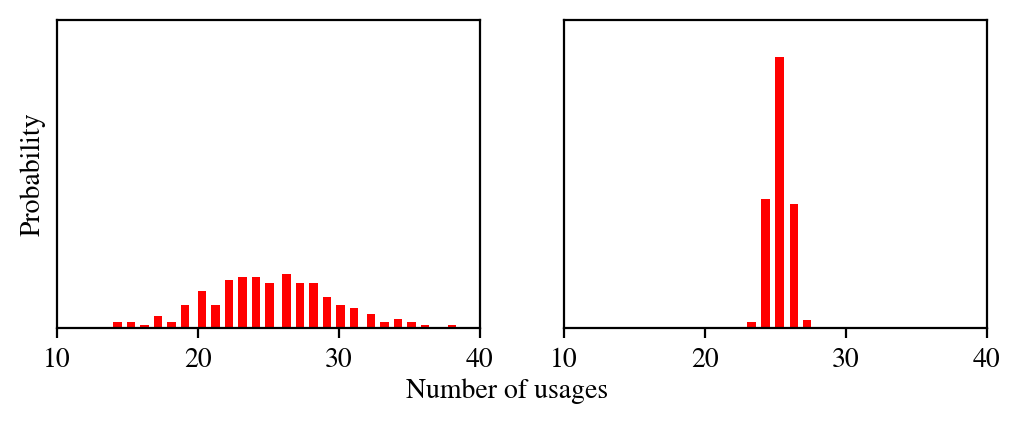}
    \end{center}
    \vspace{-15pt}
    \caption{Usage histogram of negative training images with random sampling (left) and with improved frequency-of-use sampling (right).
    Each bar represent a probability that a negative sample will be used a certain number of times.} 
    \label{fig:samp_freqs}
    \vspace{-5pt}
\end{figure}

\paragraph{Gradient-flow adjustments}

%We further propose the adjustments of the back-propagated gradient, that is required to successfully learn segmentation and classification in a single end-to-end manner. 

%We further propose two adjustments of the back-propagated gradient, that are required to successfully learn segmentation and classification in a single end-to-end manner. 

We propose to eliminate the gradient-flow from the classification network through the segmentation network, which is required to successfully learn the segmentation and the classification layers in a single end-to-end manner. First, we remove the gradient-flow through the max/avg-pooling shortcuts used by the classification network as proposed in~\cite{Tabernik2019JIM}. In Figure~\ref{fig:arch}, this is marked with the (a). Those shortcuts utilize the segmentation network's output map to speed-up the classification learning. Propagating gradients back through them would add error gradients to the segmentation's output map, however, this can be harmful since we already have error for that output in the form of pixel-level annotation. %We have experimented with enabling gradient-flow for only weakly labeled samples, where no pixel-level annotations are provided, but this has not proven useful. We therefore propose to stop the gradient-flow through max/avg-pooling shortcuts altogether.

We also propose to limit the gradients for the segmentation that originate in the classification network. In Figure~\ref{fig:arch}, this is marked with the (b).
%As the second gradient-flow adjustment, we propose to scale those gradients  
During the initial phases of the training, the segmentation net does not yet produce meaningful outputs, therefore gradients back-propagating from the classification network can negatively effect the segmentation part. We propose to completely stop those gradients, thus preventing the classification network from changing the segmentation network. This closely follows the behaviour of a two-stage learning from~\cite{Tabernik2019JIM}, where segmentation network is trained first, then the segmentation layers are frozen and only the classification network is trained in the end.

\begin{table*}
\begin{center}
\caption{Defect detection performance on DAGM dataset in terms of true positive (TPR) and true negative rates (TNR).}
\label{tab:dagm}
\begin{tabular}{cccccccccccccccccc} 
\toprule

	\multirow{3}{*}{Surface}
	& \multicolumn{2 }{c}{\textit{Ours}} 
    && \multicolumn{2 }{c}{\textit{Ra\v{c}ki et al.~\cite{Racki2018}}} 
    && \multicolumn{2 }{c}{\textit{Weimer et al.~\cite{Weimer2016}}}
    && \multicolumn{2 }{c}{\textit{Kim et al.~\cite{Kim2017c}}}
    && \multicolumn{2 }{c}{\textit{SIFT and ANN~\cite{Weimer2013a}}}
    && \multicolumn{2 }{c}{\textit{Weibull feat.~\cite{weibull}}}\\
    \cmidrule{2-3} \cmidrule{5-6} \cmidrule{8-9} \cmidrule{11-12} \cmidrule{14-15} \cmidrule{17-18} \vspace{-8pt}\\
    & TPR & TNR && TPR & TNR && TPR & TNR && TPR & TNR && TPR & TNR && TPR & TNR \\
    
\midrule
 1 & 100 & 100 && 100  & 98.8 && 100  & 100  && 99.8 & 100 && 100  & 98.9 && 98.0 & 87.0 \\
 2 & 100 & 100 && 100  & 99.8 && 97.3 & 100  && 100 & 100 && 91.3 & 95.7 && -    & - \\
 3 & 100 & 100 && 100  & 96.3 && 100  & 95.5 && 100 & 100 && 100  & 98.5 && 100  & 99.8 \\
 4 & 100 & 100 && 98.5 & 99.8 && 98.7 & 100  && 99.9 & 100 && -    & -    && -    & - \\
 5 & 100 & 100 && 100  & 100  && 100  & 98.8 && 100 & 100 && 100  & 98.2 && 100  & 97.2 \\
 6 & 100 & 100 && 100  & 100  && 99.5 & 100  && 100 & 100  && 100  & 99.8 && 100  & 94.9 \\
 7 & 100 & 100 && 100  & 100  && -    & -    && -    & -    && -    & -    && -    & - \\
 8 & 100 & 100 && 100  & 100  && -    & -    && -    & -    && -    & -    && -    & - \\
 9 & 100 & 100 && 100  & 99.9 && -    & -    && -    & -    && -    & -    && -    & - \\
10 & 100 & 100 && 100  & 100  && -    & -    && -    & -    && -    & -    && -    & - \\
\bottomrule
\vspace{-25pt}
\end{tabular}
\end{center}
  
\end{table*}

\subsection{Frequency-of-use sampling}

Current implementation of the two-stage architecture employed an alternating sampling scheme~\cite{Tabernik2019JIM} that provided balance between the positive and the negative training samples by alternating between a positive and a negative sample in each training step. We propose to improve the alternating sampling scheme by replacing the naive random sampling with one based on the frequency of use of each negative sample. The existing alternating sampling scheme forces a selection of negative images ($\Tilde{N} \subset N$) for every epoch in the same amount of positive images ($P$). However, due too $P<<N$, the selected subset $\Tilde{N}$ will be relatively small. Since current approach ~\cite{Tabernik2019JIM} employs uniform random sampling of negative images for every epoch, this leads to a significant over-use of some samples and under-use of others, as can be observed on the left side in Figure~\ref{fig:samp_freqs}.

\begin{figure}%[t!]
    \begin{center}
        \includegraphics[width=\columnwidth]{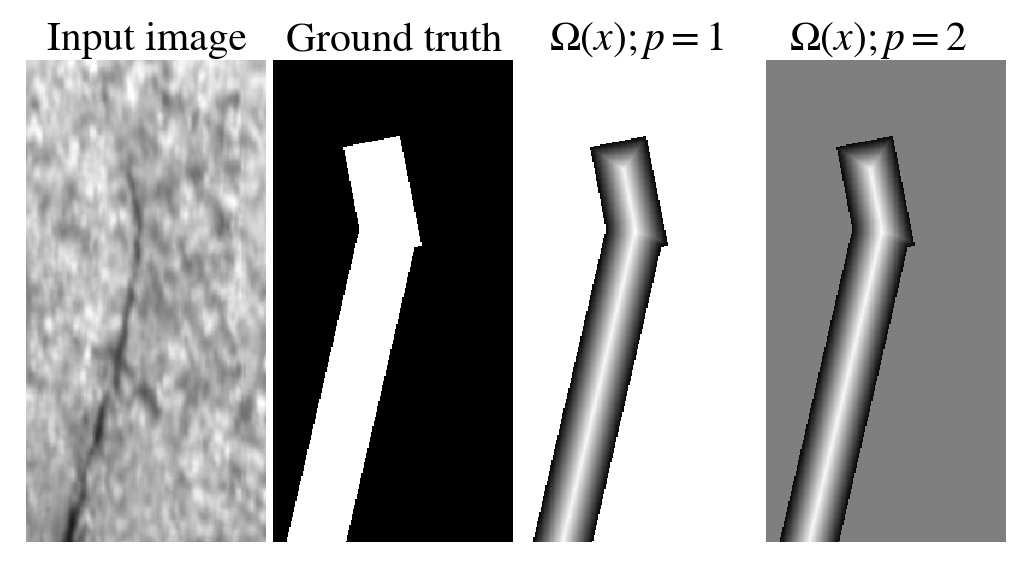}
    \end{center}
     \vspace{-15pt}
    \caption{Segmentation loss weight mask obtained by applying distance transform algorithm on the label. Whiter shades on the segmentation loss mask indicate pixels with greater weight.} 
      \vspace{-5pt}
  \label{fig:seg_loss_mask}
\end{figure}

We propose to replace the random sampling of negative examples with a one based on the frequency-of-use. We sample each image with the probability inversely proportional to the frequency of use of that image. %Histograms in Figure~\ref{fig:samp_freqs} shows the probabilities of number of appearances for each of the 200 negative training samples used during the training of a model, lasting 100 epochs, with positive-to-negative ratio of $\nicefrac{1}{4}$. 
As seen in the right histogram in Figure~\ref{fig:samp_freqs}, the frequency-of-use sampling significantly reduces the over-use and the under-use in all samples, and ensures even use of every negative image during the training process.

\subsection{Loss weighting for positive pixels}

When only approximate, region-based labels are available, such as shown in Figure~\ref{fig:seg_loss_mask}, we propose to consider the different pixels of the annotated defected regions differently. In particular, we give more importance to the center of the annotated regions and less importance to the outer parts. This alleviates the ambiguity arising at the edges of the defect, where we can not be certain whether the defect is present or not. We implement the importance in different sections of labels by weighting the segmentation loss accordingly. We weight the influence of each pixel at positive labels in accordance with its distance to the nearest negatively labelled pixel using the distance transform algorithm. 

%In order to leverage for uncertainty of annotations, we used distance transform algorithm, to obtain masks for weighting segmentation loss for each pixel. Defective pixels are transformed into their distance to nearest negative pixels, and then some function is applied over new values in order to obtain final segmentation loss mask.

We formulate weighting of the positive pixels as:
\begin{equation}\label{eq:EDT}
    %\mathcal{L}_{seg} = \frac{\sum_{\forall_{pix}} w_{pix} \cdot \mathcal{\hat L}(pix)}{\sum_{\forall_{pix}} w_{pix}},
    \mathcal{L}_{seg} = \Omega\left(\frac{\mathcal{D}(pix)}{\mathcal{D}(pix_{max})}\right) \cdot \mathcal{\hat L}(pix),
\end{equation}
where $\mathcal{\hat L}(pix)$ is the original loss of the pixel, $\nicefrac{\mathcal{D}(pix)}{\mathcal{D}(pix_{max})}$ is a  distance to the nearest negative pixel normalized by the maximum distance value within the groundtruth region and $\Omega(x)$ is a scaling function that converts the relative distance value into the weight for the loss. In general, the scaling function $\Omega(x)$ can be defined differently depending on the defect and annotation type. However, we have found that a simple polynomial function provides enough flexibility for different defect types:
\begin{equation}
    \Omega(x) = w_{pos} \cdot x^p
\end{equation}
where $p$ controls the rate of decreasing the pixel importance as it gets further away from the center, while $w_{pos}$ is an additional scalar weight for all positive pixels. We have often found $p=1$ and $p=2$ as best performing, depending on the annotation type. Examples of a segmentation mask and two weight masks are depicted in Figure~\ref{fig:seg_loss_mask}. Note, that the weights for the negatively labeled pixels remain $1$. 

\begin{table*}[]
%\footnotesize
\caption{Reported average precision (AP) on KolektorSDD dataset.}\label{tab:KolektorSDD}
\centering
%\resizebox{\textwidth}{!}{%
\begin{tabular}{lccccccc}
%\cline{2-7}
\toprule
\multirow{2}{*}{\textit{Architecture and approach}} & \multirow{2}{*}{\textit{Learning stages}} & \multicolumn{6}{c}{\textit{Number of positive training samples}} \\
\cmidrule{3-8}  \vspace{-8pt}\\
 & & \textit{33} & \textit{25} & \textit{20} & \textit{15} & \textit{10} & \textit{5}  \\
\midrule
Extended Segmentation+Decision Network (ours) & \textit{end-to-end} & \textbf{100.00} & \textbf{99.78}  & \textbf{100.00} & \textbf{99.88} & \textbf{99.31}  & \textbf{96.71} \\
Segmentation+Decision Network~\cite{Tabernik2019JIM} & \textit{separate (two stages)} & 99.0  & 97.5 & 99.5 & 97.4 & 98.8 & 95.8  \\
Cognex ViDi (commercial software)~\cite{Tabernik2019JIM} & - & 99.0 & 97.4 & 95.7 & 97.1 & 95.6 & 89.2 \\
\midrule
Xu et al.~\cite{Xu2020} (image-level label only) & \textit{\makecell[c]{separate (three stages)}} & 99.5 & - & - & - & 98.0 & -  \\
Pre-trained ResNet~\cite{Xu2020} (image-level label only) & - & 97.8 & -  &  - & - & - & - \\
\bottomrule
\vspace{-15pt}

\end{tabular}
%}
\end{table*}

\section{Experiments} \label{sec:experiments}

In this section, we extensively evaluate the proposed approach on the several anomaly detection tasks. First, we compare it against the related methods on a well-known synthetic benchmark dataset, the DAGM 2007~\cite{Weimer2016} dataset. Then we evaluate the end-to-end model with all the improvements against the baseline model with separate learning and no improvements on a fully supervised problem using the KolektorSDD~\cite{Tabernik2019JIM} dataset. We also evaluate it on the Severstal Steel Defect~\cite{SeverstalSteel2019} dataset with a larger and a more difficult set of images to demonstrate the general applicability of the proposed architecture. Finally, we assess the contributions of the individually proposed components in the ablation study.

\subsection{Performance metrics}

In all the experiments, we focus on evaluating per-image classification metric, which is the most relevant metric in the industrial quality control. In particular, we observe the average precision (AP), which is calculated as the area under the precision-recall curve and better captures the performance of the model in highly unbalanced datasets than the other established metrics, such as AUC. We also report the number of misclassifications, false positives (FP) and false negatives (FN), which are dependent on a specific threshold applied to the classification score. We report the number of misclassifications at the threshold value where the highest F-measure is achieved.

\subsection{Implementation details}

The proposed architecture and all adaptations are implemented in PyTorch. In all experiments, the network was trained with stochastic gradient descent, with no momentum and with no weight decay. Cross-entropy loss has been applied to both the segmentation and the classification networks. No image resizing and no data augmentation have been used in any of the experiments.

\subsection{The DAGM 2007 dataset}

Comparison to the related methods for the surface defect detection is performed on the DAGM 2007~\cite{Weimer2016} dataset, which is a well known benchmark database for the surface defect detection. It contains images of various surfaces with artificially generated defects. Surfaces and defects are split into 10 classes of various difficulties. In our evaluation, we consider each class as its own binary classification problem. 

The following hyper-parameters were used for all the classes. We trained for 50 epochs, with the learning rate of $\eta=0.01$ and the classification loss weight set to $\delta=1$. We used batch size of 5. Since provided labels are not always directly centered over the anomaly, we used linear scaling function $\Omega(x)$ with $p=1$. This prevented the quick drop of the weights for the defective pixels that are slightly off-centered, but still reduced the importance of pixels further away that are most likely not defective. We weighted positive samples 20-times more than negative samples  ($w_{pos}=20$). %to account for the imbalance in the number of positive and negative pixels in this dataset.

Performance in terms of true positive (TPR) and true negative (TNR) rate on all classes is shown in Table~\ref{tab:dagm}. The proposed approach achieves 100\% detection rate on all of them, thus completely solving this dataset. A  two-stage architecture with separate learning by Ra\v{c}ki et al.~\cite{Racki2018} achieved 100\% rate on half of the classes, but still had some false negatives on the other half.
A compact architecture proposed by Huang et al.~\cite{huang_sensors2020_dagm} achieved 99.79\% mAcc (mean Accuracy) over the first 6 classes. The proposed end-to-end learning and the additional improvements increase this result even further and achieves 100\% detection rate on the remaining surfaces as well.
Few examples of the images, defects and segmentation network outputs are shown in Fig.~\ref{fig:dagm_ksdd_true}.

\subsection{The KolektorSDD }

Next, we evaluate the proposed approach on KolektorSDD~\cite{Tabernik2019JIM} to compare it against the existing two-stage state-of-the-art architecture. We followed the same evaluation protocol as described in~\cite{Tabernik2019JIM}, using a 3-fold cross-validation with the same train-test split. In contrast to~\cite{Tabernik2019JIM}, we did not apply exponential moving average to the network's weights during the learning. When reporting performance metrics for cross-validation, we avoid combining uncalibrated score from the individual folds, but instead calculate the metric for each individual fold and report an average over all three folds for AP, while for FP and FN, we report the sum of false values over all three folds. 

We used a rotated bounding-box annotation weighted by the distance transform since this annotation requires less manual work to be obtained. For scaling function $\Omega(x)$ applied to the annotations, we used $w_{pos}=1$ and $p=2$ since KolektorSDD contains mostly thin surface fractures that are fairly prominent at the center of the annotation.
%For scaling function $\Omega(x)$ parameters applied to the annotation, we used $w_{pos}=1$ and $p=2$. Since KolektorSDD contains thin surface fractures that are mostly prominent at the center of the annotation, we used square function that quickly decreases the weight factor further away from the center of the defect.

The learning rate was $\eta=0.5$, while the classification loss weight  was set to $\delta=0.01$. When training with all samples the network was trained for 35 epochs, while in the experiments with smaller numbers of positive samples, the number of steps was kept approximately the same and due to a smaller training set size resulted in training for 40, 50, 70, 90 and 150 epochs for training sets with 25, 20, 15, 10 and 5 positive samples, respectively. 

\begin{figure*}%[h]
    \vspace{-10pt}

    \begin{center}
    \includegraphics[width=0.95\linewidth]{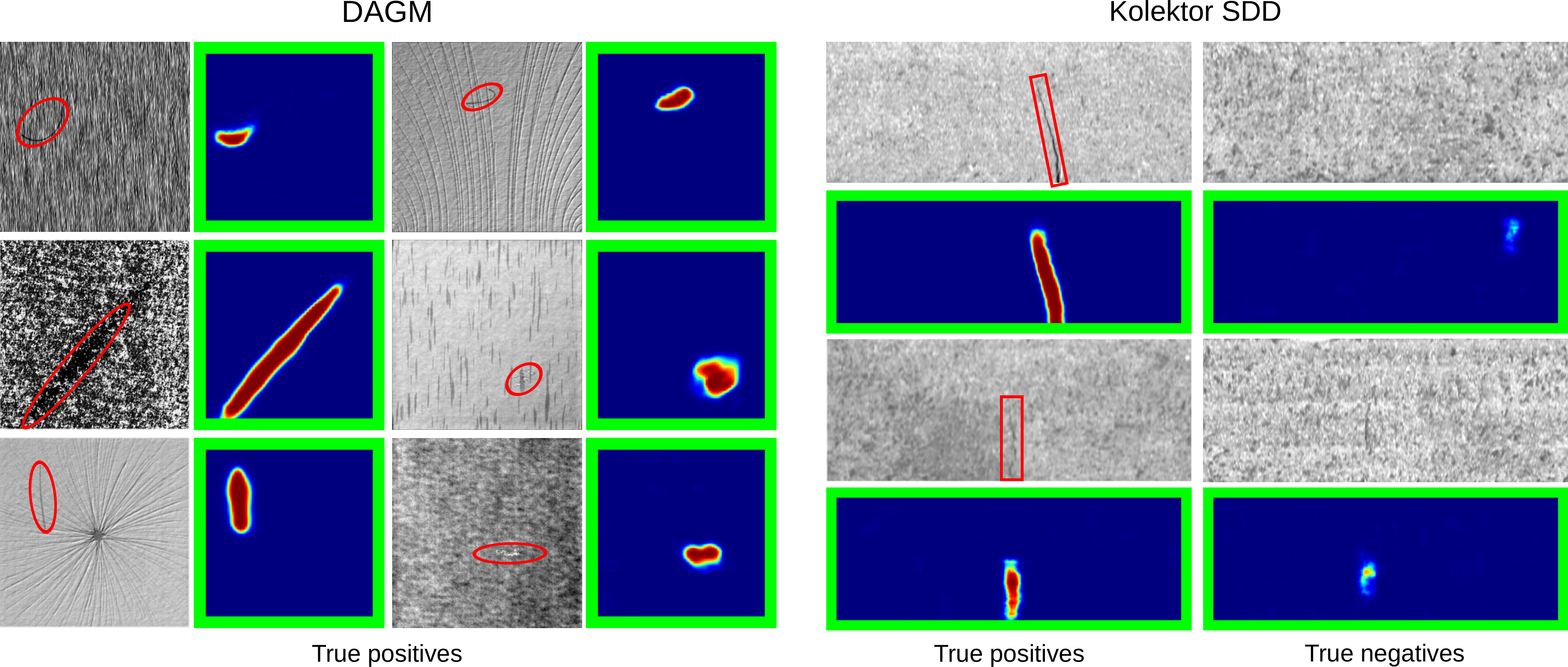}
    \end{center}
    \vspace{-10pt}
    \caption{Examples of images, defects and detections with segmentation output from the DAGM (left) and the KolektorSDD (right) datasets.} 
    \vspace{-10pt}
    \label{fig:dagm_ksdd_true}
\end{figure*}

Results are reported in Table~\ref{tab:KolektorSDD} and show the average precision (AP) for different number of positive samples used. As a baseline we use the two-stage architecture~\cite{Tabernik2019JIM} trained with the same rotated bounding-box annotation but with two-stage learning procedure instead of the end-to-end learning and without our improvements. Our proposed method out-performs the baseline in all cases as seen in Table~\ref{tab:KolektorSDD}. This is shown across different numbers of positive training samples as end-to-end learning outperforms the baseline by around 1-2\% in most cases. Moreover, our approach achieved AP of 100\% with no misclassifications when using all positive training samples. The proposed approach also outperforms all other related methods, including the commercial software that was trained with even more precise annotations, and a two-stage architecture trained with weak supervision~\cite{Xu2020}. Although, Xu et al.~\cite{Xu2020} used only image-level labels, our approach with some pixel-level annotation presents a better trade-off between no annotation and precise annotation without compromising the overall performance.
Fig.~\ref{fig:dagm_ksdd_true} presents some examples of the images, defects and detections with outputs from the segmentation network.

%the and the commercial software reported in~\cite{Tabernik2019JIM}, the proposed simultaneous learning achieves better results in all cases. With all positive training samples used in learning, our end-to-end method achieves AP of 100\% with no miss-classifications, while with only 10 positive samples it still achieves AP of 99.0\%, which is by 2.3 percentage points better than the separate learning scheme.

\begin{table*}[]
\caption{Performance of individual components on all three dataset by disabling each one.}
\label{tab:ablation_1}
\centering
\begin{tabular}{cccccccccccc}
\toprule
    \multicolumn{2}{c}{\textit{DAGM}} && \multicolumn{2}{c}{\textit{KolektorSDD}} && \multicolumn{2}{c}{\textit{Severstal Steel}} & \multirow{2}{*}{\textit{\makecell[c]{Dynamically\\balanced loss} } } & \multirow{2}{*}{\textit{\makecell[c]{Gradient-flow\\adjustment}}} & \multirow{2}{*}{\textit{\makecell[c]{Frequency-of-use\\sampling} } } &  \multirow{2}{*}{\textit{\makecell[c]{Distance\\transform} } } \\
    \cmidrule{1-2} \cmidrule{4-5} \cmidrule{7-8} \vspace{-8pt}\\
    AP & FP+FN && AP & FP+FN && AP & FP+FN &  &  &  & \\ 
\midrule
    99.95 & 2+2 && 99.75 & 1+0 && 98.32 & 73+64 & & \checkmark & \checkmark & \checkmark \\ 
    91.88 & 471+68 && 99.88 & 1+0 && 97.80 & 57+60 & \checkmark &  & \checkmark & \checkmark \\ 
    99.996 & 0+1 && 99.77 & 0+1 && 98.54 & 36+75 & \checkmark & \checkmark & &  \checkmark \\ 
    100.00 & 0+0 && 99.88 & 1+0 && 98.24 & 52+58 & \checkmark & \checkmark &  \checkmark &  \\ 
    100.00 & 0+0 && 100.00 & 0+0 && 98.74 & 59+40 & \checkmark & \checkmark & \checkmark & \checkmark \\ 
\bottomrule     
\vspace{-15pt}

\end{tabular}
\end{table*}

\subsection{The Severstal Steel defect dataset }

\begin{figure}%[h]
    \begin{center}
    \includegraphics[width=0.89\columnwidth]{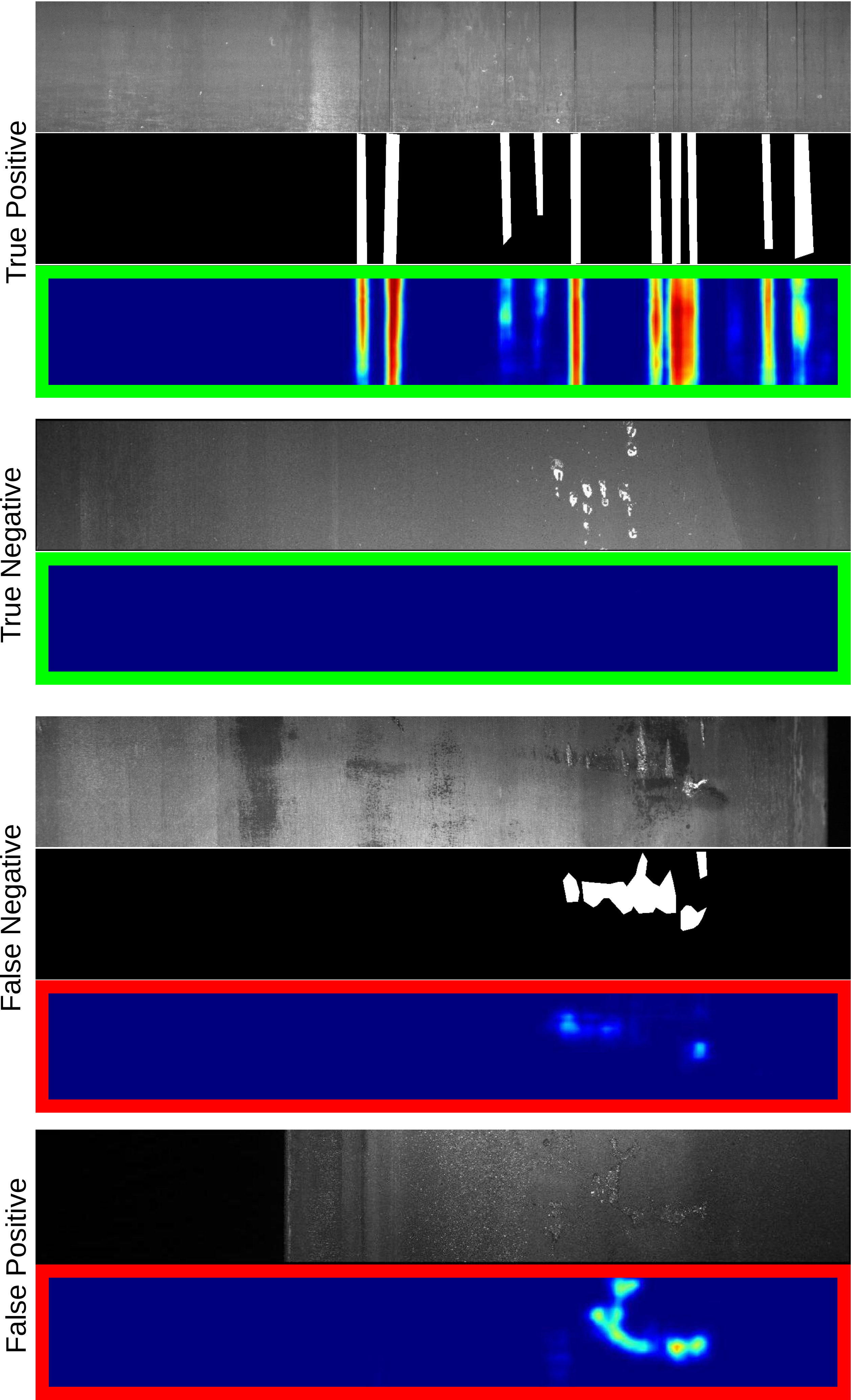}
    \end{center}
    \caption{Examples of images, defects and detections with segmentation output from the Severstal Steel defect dataset.} 
    \vspace{-15pt}
    \label{fig:steel_tp}
\end{figure}

Results on the previous two dataset are shown to be saturated as both datasets contain too few examples that would be difficult for the deep-learning approaches. We therefore additionally evaluate on the Severstal Steel defect~\cite{SeverstalSteel2019} dataset that contains a larger set of more difficult samples. The dataset consists of 12,568 images in total, with 4 different types of defects. We used a subset of images for our evaluation. In particular, we used all negative images and all positive images with only \textit{class 3} defect present, which is the most common defect in this dataset. Images that also had other types of defects were not used. This resulted in 6,666 negative images and 4,759 positive images, which were further split into the train, the validation and the test set. For training, we used 3,000 positive and 4,143 negative images, for validation, 559 positive and negative images, and for test, 1,200 positive and negative images were used. Few images from the dataset, alongside the detections with the outputs from the segmentation network are presented in Fig~\ref{fig:steel_tp}. Existing benchmarks on Severstal Steel dataset are performed with a per-pixel basis evaluation, however, we instead report metric on a per-image bases, since per-pixel accuracy of segmentation is not crucial for practical application of defect detection. Instead, it is more important that images are correctly classified as defective or non-defective.% than misclassifying several pixels around the edges of the defect.

\begin{table}[]
\caption{Average precision and misclassifications for the Severstal Steel Defect dataset using our proposed approach.}
\label{tab:steel_results}
\centering
\begin{tabular}{lcccc}
\toprule
    \multirow{3}{*}{\textit{Metric} } & \multicolumn{4}{c}{\textit{Number of positive training samples } } \\
    \cmidrule{2-5} \vspace{-8pt}\\
    & \textit{3000} & \textit{1500} & \textit{750} & \textit{300} \\
\midrule
    %\multirow{3}{*}{\makecell[c]{Extended Seg.+Dec. Net\\ with end-to-end\\learning (ours)}} 
    \textit{Average precision (AP)} & 99.04 & 99.00 & 98.91 &  97.78 \\
    \textit{False positives (FP)} & 34 & 41 & 52 & 95 \\
    \textit{False negatives (FN)} & 54 & 70 & 65 & 77 \\
\bottomrule     
\vspace{-25pt}
\end{tabular}
\end{table}

The results are shown in Table~\ref{tab:steel_results}. The positive training samples size was also varied between 300, 750, 1500 and 3000 samples. We trained for 90, 70, 50 and 40 epochs, when using 300, 750, 1500 and 3000 positive samples, respectively, and then used the best model, according to the AP on the validation set, for evaluation. The learning rate was set to $\eta=0.1$, the classification loss weight  was set to $\delta=0.1$ and we used $p=2$ for weighting positive pixels. We used the batch size of 10. Results show that with 300 positive training samples the proposed approach is able to achieve AP of $97.78$\% while with all 3000 training samples the AP increases to $99.04$\% and the number of misclassifications almost halves. Despite high AP value this still leaves 34 false positives and 54 false negatives. Several examples of detections that well demonstrate the difficulty of this dataset are depicted in Figure~\ref{fig:steel_tp}.

\begin{table*}[]
\caption{Performance of individual components on all three datasets by gradually including each one.}
\label{tab:ablation_2}
\centering
\begin{tabular}{ccccccccccccc}
\toprule
    \multicolumn{2}{c}{\textit{DAGM}} && \multicolumn{2}{c}{\textit{KolektorSDD}} && \multicolumn{2}{c}{\textit{Severstal Steel}} & \multirow{2}{*}{\textit{\makecell[c]{Dynamically\\balanced loss} } } & \multirow{2}{*}{\textit{\makecell[c]{Gradient-flow\\adjustment}}} & \multirow{2}{*}{\textit{\makecell[c]{Frequency-of-use\\sampling} } } &  \multirow{2}{*}{\textit{\makecell[c]{Distance\\transform} } } \\
    \cmidrule{1-2} \cmidrule{4-5} \cmidrule{7-8} \vspace{-8pt}\\
    AP & FP+FN && AP & FP+FN && AP & FP+FN &  &  &  & \\ 
\midrule
    90.84 & 661+45 && 99.77 & 0+1 && 95.90 & 59+102 &  &  &  & \\ 
    97.60 & 26+24 && 99.88 & 1+0  && 97.43 & 76+72 & \checkmark &  &  &  \\ 
    99.998 & 1+0 && 99.90 & 1+0 && 97.59 & 65+61  & \checkmark & \checkmark &  &  \\ 
    %100.00 & 0 && 99.54 & && 98.58 & & \checkmark & \checkmark &  & \checkmark \\ 
    100.00 & 0+0 && 99.88 & 1+0 && 98.24 & 52+58 & \checkmark & \checkmark &  \checkmark &  \\ 
    100.00 &  0+0 && 100.00 & 0+0 && 98.74 & 59+40 & \checkmark & \checkmark & \checkmark & \checkmark \\ 
\bottomrule     
\vspace{-15pt}

\end{tabular}
\end{table*}

%\subsection{Contributions of individual components}
\subsection{Ablation study}
\label{sec:ablation-stduy}

Finally, we evaluate the impact of individual components that each contributes to the results achieved above. We preform the ablation study on all three datasets, KolektorSDD, DAGM and Severstal Steel Defect dataset, however, for the Severstal Steel dataset we used only a subset of images to reduce the training time. We used 1,000 positive and negative samples for training and the same amount for testing.

We report two experiments that show the contributions of the individual components. In both experiments the identical settings and hyper-parameters are used as the ones reported before. In the first experiment, we report the performance by disabling only the specific component, while leaving all remaining ones enabled. Results are reported in Table~\ref{tab:ablation_1}. The second experiment is reported in Table~\ref{tab:ablation_2}, which demonstrates the performance when the individual components are gradually enabled, including the performance with none of the components enabled at all. Results well show that the worst performance is achieved with none of the components enabled on all three datasets, while the best results on all three are achieved only when all four components are enabled. We describe the contribution of each component to the overall improvements in more details below.

\paragraph{Dynamically balanced loss and gradient-flow adjustment}

Enabling only dynamic balancing of loss with gradual inclusion of classification network already provides a significant boost of the performance to all three datasets. Without dynamically balanced loss, i.e., without using any $\lambda$, the AP is 90.84\%, 99.77\% and 95.90\% for DAGM, KolektorSDD and Severstal Steel datasets, respectively. However, with $\lambda$ dynamically balancing the loss over the training epochs, the AP increases to 97.60\%, 99.88\% and 97.43\%, for DAGM, KolektorSDD and Severstal Steel datasets, respectively, as shown in Table~\ref{tab:ablation_2}.

Replacing dynamically balanced loss with only the gradient-flow adjustment, i.e., stopping the gradients to segmentation, would result in similar performance on the KolektorSDD and the Severstal Steel datasets, as can be observed in the first two rows in Table~\ref{tab:ablation_1}. Since both methods effectively prevent the unstable segmentation features to significantly affect the learning of the classification layers in the early stages of the learning, they naturally result in a similar performance. 
%However, using either method on the DAGM dataset is not sufficient as in both cases we experienced difficulty during the learning process where the model would not converge at all without further decreasing the classification learning rate factor to $\delta=0.01$. A more robust solution is to use both gradient-flow adjustment and dynamically balanced loss at the same time while leaving the $\delta=0.1$ that produces the much better results.
However, using either method on the DAGM dataset is not sufficient as in both cases we experienced difficulty during the learning process where the model would not converge at all without further decreasing the learning rate to $\eta=0.01$. A more robust solution is to use both the gradient-flow adjustment and the dynamically balanced loss at the same time while leaving the $\eta=0.1$ which produces much better results.
This eliminates the convergence issues while also significantly improving the results for the DAGM dataset, where 99.998\% detection rate with one misclassification can be achieved. Performance on the other two datasets also increases slightly when both components are used.

The above results demonstrate that using only the dynamically balanced loss or only the gradient-flow adjustment is not sufficient to introduce the end-to-end learning. Instead, both improvements are required to achieve the convergence and good results over the different defect detection problems. 

\paragraph{Frequency-of-use sampling}

Replacing naive random sampling in alternating scheme with one based on the frequency-of-use further improves the results. As shown in Table~\ref{tab:ablation_2}, enabling frequency-of-use sampling with previous two components improves performance from 97.59\% to 98.24\% AP on the Severstal Steel dataset, and from 99.998\% to 100\% on the DAGM dataset. On the KolektorSDD, the frequency-of-use sampling did not have significant effect as the number of misclassifications remained the same, despite the minor decrease in AP. 

\paragraph{Loss weighting for positive pixels}

Lastly, enabling the distance transform as the weights for the positive pixels pushes performance on KolektorSDD to 100\% detection rate, therefore completely solving KolektorSDD and DAGM dataset. Loss weighting also improves result on Severstal Steel dataset, increasing the AP from 98.24\% to 98.70\% as shown in Table~\ref{tab:ablation_2}. Moreover, as observed in Table~\ref{tab:ablation_1}, where only a specific component is disabled, the frequency-of-use sampling does show to improve result on the KolektorSDD when also taking into the consideration the distance transform. Result improve from 99.77\% AP and one misclassification when frequency-of-use sampling is not used to 100\% detection rate when it is. This indicates that a combination of the distance transform and the frequency-of-use sampling works the best across the different datasets.

\section{Conclusion} \label{sec:conclusion}

In this paper, we presented a simultaneous learning scheme for a two-stage deep learning architecture used in the surface anomaly detection. We proposed an end-to-end mechanism that facilitates simultaneous learning by dynamically balancing the loss between the learning of the segmentation and the classification networks, as well as through accordingly adjusting the gradient-flow from the classification to the segmentation layers. Although training of the segmentation layers is not affected by the errors originating in the classification layers, we still consider this approach an end-to-end learning since the training is performed in a unified network with one forward and one backward pass. We additionally proposed to extend the loss function to account for uncertainty of region based annotations, thus allowing to use the less precise pixel-level annotations that are easier to obtain than the high-precision labels. Such labels are useful for many different cases in thenindustrial defect detection where pixel-wise location of defects are not always easily defined, such as in KolektorSDD~\cite{Tabernik2019JIM} and in several classes of the DAGM~\cite{Weimer2016} dataset. In such cases, a more broad region can easily encompass the defective pixels and thus can take much less effort to annotate. Finally, we also introduced the frequency-of-use sampling of defected samples to ensure the even use of the negative samples in the unbalanced datasets.

We demonstrated the benefits of the proposed approach on three defect detection problems. In particular, 100\% detection rate has been achieved on the DAGM~\cite{Weimer2016} and the KolektorSDD~\cite{Tabernik2019JIM} datasets, thus out-performing all state-of-the-art methods and completely solving those problems. On the KolektorSDD we also demonstrated that end-to-end scheme can successfully replace the two-stage learning process from~\cite{Tabernik2019JIM}, leading to the halving of the training iterations needed, while even improving the classification accuracy. Since the DAGM and the KolektorSDD are now solved, we also demonstrated the performance on the Severstal Steel defect~\cite{SeverstalSteel2019} dataset, which includes large number of more difficult defects than the previous two datasets. Due to the size and the difficulty of this dataset, we hope that the results will serve as a future benchmark model to compare the performance of the new methods for the surface defect detection.

Finally, the extensive ablation study demonstrated the behaviour of individually proposed improvements. Each proposed component has proven important and beneficial since removing any one decreased the accuracy. The study also demonstrated that to correctly implement end-to-end learning both dynamically balanced loss and gradient-flow adjustment between the segmentation and the classification networks are needed. Without them, the results were either worse or the learning had difficulties converging. Study has also shown frequency-of-use sampling and distance transform as both individually important in improving results for specific datasets. However, the combination of both has proven much better, thus making the proposed improvements mutually beneficial. This significantly contributed to achieving state-of-the-art results across three different defect detection problems.

\section*{Acknowledgements}
 This work was in part supported by the ARRS research project J2-9433 (DIVID) and research programme P2-0214.
 
{%\small
\bibliographystyle{ieee}
\bibliography{library}
}

\end{document}